%% file: smic.tex
\documentclass[a4paper,fleqn]{cas-sc}

\usepackage[authoryear,longnamesfirst]{natbib}
\input{math_commands.tex}

\usepackage{graphicx}
\usepackage{hyperref}
\usepackage{url}
\usepackage{algorithm}
\usepackage{algorithmic}
\usepackage{amsfonts}
\usepackage{microtype}
\usepackage{xcolor}
\usepackage{amsmath}
\usepackage{amsthm}
\usepackage{amssymb}
\usepackage{array}
\usepackage{booktabs}
\usepackage{newfloat}
\usepackage{listings}
\usepackage{subcaption}
\newtheorem{proposition}{Proposition}

\theoremstyle{remark}
\newtheorem{remark}{Remark}
\theoremstyle{definition}

\def\tsc#1{\csdef{#1}{\textsc{\lowercase{#1}}\xspace}}
\tsc{WGM}
\tsc{QE}

\begin{document}
\let\WriteBookmarks\relax
\def\floatpagepagefraction{1}
\def\textpagefraction{.001}

\shorttitle{SMIC for Natural Language Inference}

\shortauthors{X. Zhang and V.S. Sheng}

\title [mode = title]{SMIC: Shared Multiscale Input-Reference Conditioning across Transformer Layers for Natural Language Inference}
%

\author[1]{Xin Zhang}[orcid=0009-0009-8935-0720]
\cormark[1]
\ead{xin.zhang@ttu.edu}
\credit{Conceptualization, Methodology, Software, Investigation, Formal analysis, Visualization, Writing -- original draft}

\author[1]{Victor S. Sheng}[orcid=0000-0003-4960-174X]
\ead{victor.sheng@ttu.edu}
\credit{Supervision, Writing -- review \& editing}

\affiliation[1]{organization={Department of Computer Science, Texas Tech University},
	addressline={Box 43104},
	city={Lubbock},
	postcode={79409-3104},
	state={TX},
	country={USA}}

\cortext[1]{Corresponding author}


\begin{abstract}
Natural Language Inference models can be sensitive to localized lexical changes. We introduce Shared Multiscale Input-Reference Conditioning (SMIC), which constructs a single input-dependent multiscale reference and reuses it across Transformer layers. Each layer applies its own multiscale transformation to the shared reference and injects the resulting conditioning feature before the attention update. We instantiate SMIC in a token--context Transformer (TCT) and conduct from-scratch comparisons against decoder-based and Multi-Layer Perceptron (MLP) baselines across all three ANLI rounds. TCT-SMIC attains the highest observed ten-seed mean accuracy on all three ANLI rounds. The ablations further support the contribution of SMIC. We evaluate a real-valued Llama-style plug-in and extend deletion/insertion diagnostics to pretrained ModernBERT encoder and Llama decoder models with and without SMIC. Across the evaluated perturbation protocols, TCT-SMIC exhibits comparatively low pooled-attention redistribution and smaller signed accuracy changes. We further establish a finite-sample bound on average subset-mass redistribution in terms of the mean Jensen--Shannon divergence between the restricted and renormalized attention distributions.
\end{abstract}
\maketitle

\section{Introduction}\label{sec:introduction}

Scaling Transformer-based language models increases training and inference costs, while larger parameter counts do not always produce proportional gains under comparable optimization budgets \citep{kaplan2020scaling,yao2022nlp,petty2024impact}. Meanwhile, Natural Language Inference (NLI) remains challenging, particularly on ANLI, an adversarial benchmark requiring models to distinguish entailment, contradiction, and neutrality \citep{nie2019adversarial}. Previous work has also shown that NLI predictions can be sensitive to localized lexical changes \citep{glockner-etal-2018-breaking,naik2018stress}. This motivates architectures that process localized lexical information without relying solely on larger models.

A sparse lexical replacement changes only a few input tokens, yet it can change the predicted relation. Some replacements should change the relation and others should not. We therefore treat perturbation sensitivity as a diagnostic of model behavior.

Some NLI encoders let higher layers read the input embeddings again \citep{nie2017shortcut,talman2019sentence,kim2019semantic}, and looped Transformers add the input back at every iteration \citep{yang2024looped}. These designs reuse the input at the level of single tokens. Defenses against word substitutions change the training data, the training objective, the input, or the inference procedure \citep{dong2021towards,zhou-etal-2021-defense,wang2023rmlm}, and certified defenses guarantee an unchanged prediction for a defined set of edits \citep{jia-etal-2019-certified,zeng2023certified}. None of these studies builds multiscale features from the original input once and conditions every Transformer layer on them through a separate transformation per layer. Section~\ref{sec:relatedwork} reviews these lines of work.

\begin{figure*}[t]
	\centering
	\includegraphics[width=0.85\linewidth]{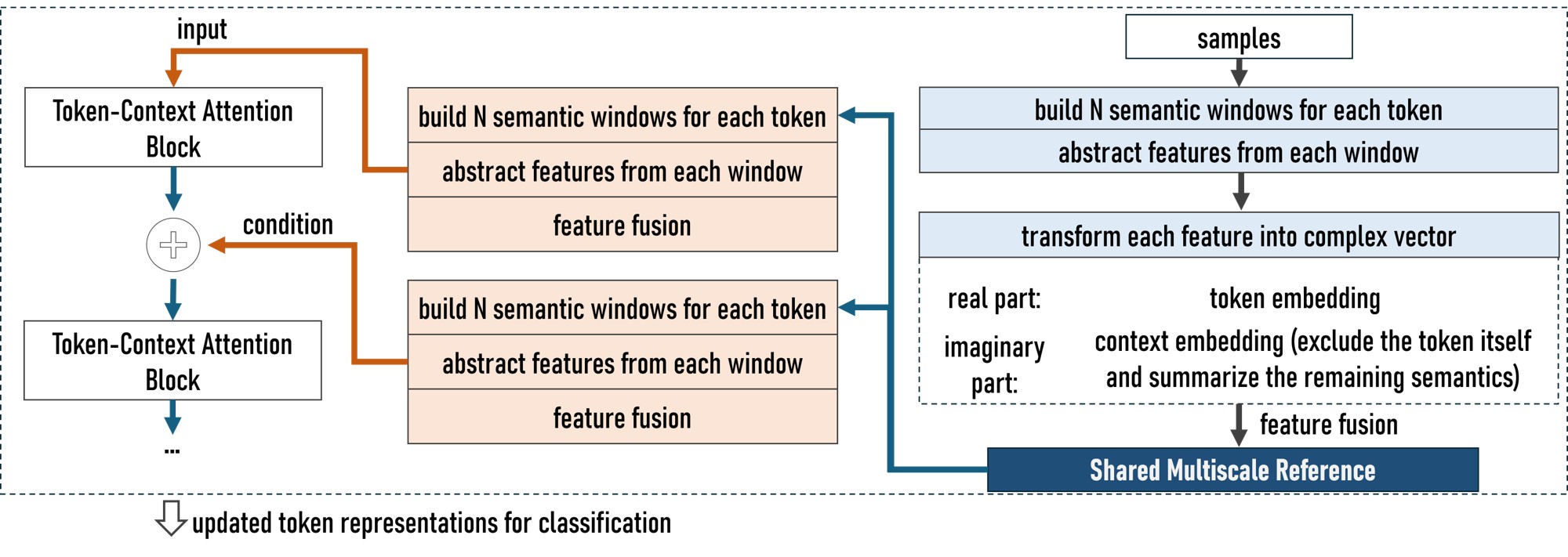}
	\caption{Overview of TCT-SMIC. The model builds one multiscale reference from the premise and hypothesis embeddings and shares it across all attention layers. Each layer transforms the reference with its own convolutions and adds the result to its token and context components before Token--Context Attention.}
	\label{fig:whole}
\end{figure*}

Inspired by multiscale reference methods in signal processing \citep{martin2001noise,cohen2003noise,wang2017novel}, we introduce Shared Multiscale Input-Reference Conditioning (SMIC), which constructs an input-dependent reference from multiscale convolutional features and gives every layer direct access to it. Building on \citet{2026tokencontext}, our token--context Transformer with SMIC (TCT-SMIC) maps scale-specific features to complex--vector token representations and fuses them into the shared reference. Each layer applies its own multiscale transformation to this reference and injects the resulting features into the token and context components before the Token--Context Attention (TCA) update. Figure~\ref{fig:whole} summarizes the TCT-SMIC architecture.

\paragraph{Contributions.}

\textbf{First}, we introduce SMIC, an architecture-level method that builds one multiscale reference from the input and conditions every layer on it. SMIC aims to improve ANLI accuracy and reduce sensitivity to sparse lexical replacements without relying solely on larger models.
\textbf{Second}, we show what the two diagnostics measure. Signed accuracy change equals the break rate minus the recovery rate, and the mean Jensen--Shannon divergence bounds how much attention any chosen set of tokens can gain or lose on average. Both results hold for any model, so all configurations are read on the same scale.
\textbf{Third}, TCT-SMIC reaches the highest ten-seed mean accuracy among decoder-based and MLP baselines of similar size on all three ANLI rounds, with lower training FLOPs than most baselines in each round. Ablations compare the multiscale reference with a token-level input reference, a variant without learned span scales, and a layer-local reference. Under sparse lexical replacements, TCT-SMIC shows comparatively low attention redistribution and the smallest mean signed accuracy change. A real-valued Llama-style plug-in and deletion/insertion diagnostics on pretrained ModernBERT and Llama backbones evaluate SMIC across real-valued and pretrained architectures.

\section{Related work}\label{sec:relatedwork}

\paragraph{Lexical sensitivity in NLI.}
Several studies show that NLI predictions can depend on a few words of the input. A test set whose hypotheses differ from their premises by a single word substantially lowers the accuracy of models trained on a standard NLI dataset \citep{glockner-etal-2018-breaking}. HANS shows that models trained on a large NLI dataset adopt heuristics such as word overlap between the premise and the hypothesis, and these models perform poorly on examples where the heuristics fail \citep{mccoy2019right}. Stress tests that append tautologies to the input or introduce spelling errors reveal further weaknesses of sentence-encoder NLI models \citep{naik2018stress}. ANLI was collected over three rounds from hypotheses that fooled a model in the loop \citep{nie2019adversarial}. Asking a language model to explain its label improves its accuracy on ANLI and makes single words less predictive of the label \citep{kavumba2023prompting}. KACE edits key phrases of the input to build counterfactual examples and uses them to generate contrastive explanations for NLI \citep{chen2021kace}. These studies evaluate NLI models or change how the models are prompted and explained. We change the model architecture and measure how its accuracy and attention respond to sparse lexical replacements on ANLI.

\paragraph{Reusing the input across layers.}
Some NLI encoders give higher layers direct access to the input. Shortcut-stacked sentence encoders feed the word embeddings and the outputs of all lower layers to each recurrent layer \citep{nie2017shortcut}. Iterative refinement encoders let every recurrent layer read the input embeddings again \citep{talman2019sentence}, and densely connected co-attentive networks pass the features of all preceding layers upward \citep{kim2019semantic}. Looped Transformers add the original input back at every iteration, a technique known as input injection \citep{yang2024looped}. Deep equilibrium models also inject the input at every step of their fixed-point iteration \citep{bai2019deep}. In Transformer language models, value residual learning adds the values of the first attention layer to those of every later layer \citep{zhou-etal-2025-value}, and ExoFormer blends queries, keys, and values computed from the input embeddings into those of every layer \citep{su2026attention}. These designs reuse the token embeddings or the hidden states of earlier layers. Feature-wise linear modulation instead scales and shifts intermediate features with parameters computed from a separate conditioning input \citep{perez2018film}. SMIC conditions every layer on the input itself. It builds multiscale span features from the input once, and each layer transforms them with its own convolutions before adding them to the layer input. Section~\ref{sec:abla} tests this difference by replacing the multiscale reference with one computed directly from token embeddings.

\paragraph{Convolution and multiscale features.}
Convolutional sentence models apply filters of several widths to word embeddings \citep{kim-2014-convolutional}, and stacked convolutions have been used for language modeling and machine translation \citep{dauphin2017language,gehring2017convolutional}. Some Transformer variants add locality inside each layer. The Multi-Scale Transformer gives its attention heads windows of different sizes \citep{guo2020multi}. Lite Transformer splits each layer into a convolution branch for local context and an attention branch for long-range context \citep{wu2020lite}. In these models, the convolutions or windowed attention of each layer operate on the output of the layer below. SMIC applies its convolutions to a reference built from the input, so every layer receives multiscale features of the input itself.

\paragraph{Certified robustness to word substitutions.}
Certified defenses guarantee that a prediction does not change under any edit in a defined set. Early methods certify robustness to synonym substitutions with interval bound propagation \citep{jia-etal-2019-certified} or with differential privacy \citep{wang2021certified}. Randomized smoothing certifies a prediction by aggregating the predictions of a model over random changes to the input. These changes include masking words \citep{zeng2023certified}, adding Gaussian noise in a learned semantic space \citep{zhao2022certified}, and perturbing word embeddings continuously \citep{zhang2024random}. Text-CRS extends certification beyond substitutions to word reordering, insertion, and deletion \citep{zhang2024text}. RS-Del certifies sequence classifiers against bounded edit distance through random deletion \citep{huang2023rs}, and UniT unifies certified robust training against text perturbations \citep{ye2023unit}. SMIC does not produce certificates. It changes the model architecture, and we measure its sensitivity empirically.

\paragraph{Empirical defenses against word substitutions.}
Empirical defenses improve accuracy under attack without a formal guarantee. ASCC-defense models the substitutions of each word as the convex hull of their embeddings and trains on the worst case within this hull \citep{dong2021towards}. Dirichlet neighborhood ensembles sample virtual inputs from the convex hull of each word and its synonyms and train on them \citep{zhou-etal-2021-defense}. Synonym encoding maps each cluster of synonyms to one code before the input layer \citep{wang2021natural}. Self-supervised contrastive learning with adversarial perturbations improves robustness without labeled data \citep{meng2022self}, and synonym-unaware fast adversarial training perturbs the embedding space without a synonym list \citep{yang2025synonym}. RMLM mainly acts at inference time. It randomly replaces words with synonyms, corrects the context with a BERT defender, and filters out adversarial examples with a detector \citep{wang2023rmlm}. These defenses change the training data, the training objective, the input, or the inference procedure. SMIC changes only the forward computation of the model.

\paragraph{Complex-valued representations.}
Complex-valued neural networks process features that carry both an amplitude and a phase \citep{lee2022complex}. EulerFormer turns token embeddings into polar-form complex vectors and models semantic and positional differences as rotations \citep{tian2024eulerformer}. The complex--vector token representation (CVTR) assigns token and context roles to the real and imaginary parts of each token representation \citep{2026tokencontext}. TCT-SMIC builds on CVTR and TCA. Sections~\ref{sec:attnfaith} and~\ref{sec:work-pertur} also add SMIC to a real-valued Llama-style model to test it without CVTR and TCA.

\section{Preliminaries}\label{sec:background}

We build on the complex--vector token representation (CVTR) and Token--Context Attention (TCA) introduced by \citet{2026tokencontext}. For each embedding coordinate, CVTR computes a sequence-level magnitude and assigns token-dependent phases that encode each token's relation to this magnitude. Given token embeddings $w_1,\ldots,w_n\in\mathbb{R}^{d}$, conversion to Cartesian coordinates yields $Z_j=T_j+\mathrm{i}C_j\in\mathbb{C}^{d}$, where $T_j=\Re(Z_j)$ and $C_j=\Im(Z_j)$ are the coupled real-valued token and context components, respectively. Appendix~\ref{app:complex-derivation} provides the Cartesian-form derivation.

\section{Method}
\label{sec:tct}

TCT-SMIC works in three steps (Figure~\ref{fig:whole}). It first builds one complex-valued reference for each premise--hypothesis pair from convolutional features of the input at several span sizes (Section~\ref{sec:tct_reference}). The reference is computed once and shared by all attention layers. Each layer then applies its own multiscale convolutions to the reference and adds the weighted result to its token and context components (Section~\ref{sec:tct_residual}). Token--Context Attention finally updates the conditioned components (Section~\ref{sec:tct_update}). The reference depends only on the input, whereas its transformations and injection weights differ across layers. Appendix~\ref{app:settings} lists all fixed settings.

\subsection{Shared Multiscale Input Reference}
\label{sec:tct_reference}

Let $u\in\{p,h\}$ denote the premise and hypothesis. For each $u$, let $E^{u}\in\mathbb{R}^{n_u^{\mathrm{pad}}\times d}$ denote the padded embedding tensor, let $\chi^{u}\in\{0,1\}^{n_u^{\mathrm{pad}}}$ denote the active-position mask, and let
$n_u=\sum_{j=1}^{n_u^{\mathrm{pad}}}\chi_j^u$
be the active length. We project the two sequences separately:
$\widetilde{E}_{j}^{u}=\chi_j^u\left(W_{u}E_{j}^{u}+b_{u}\right).$
The premise branch uses the candidate kernel set
$
\mathcal{K}_{p}
=
\{1,3,7,13,21,33\},
$
whereas the hypothesis branch uses
$
\mathcal{K}_{h}
=
\{1,3,5,7,9,13\}.
$
Natural-language dependencies are generally local. Related words tend to occur at shorter linear distances \citep{futrell2015large}. We therefore use denser coverage of short receptive fields while retaining larger kernels for long-range interactions. Because premises have a larger maximum length, their kernel bank covers a wider range than the hypothesis bank. Kernel sizes are fixed, while the filters, effective centers, and mixture weights are learned jointly with the model.

For each $\kappa\in\mathcal{K}_{u}$, SMIC computes the length-preserving feature
$Y_{u,\kappa}=\operatorname{ReLU}\left(\operatorname{Conv}_{u,\kappa}\left(\widetilde{E}^{u}\right)\right).$
The trainable $\kappa=1$ convolution is initialized as an identity mapping and forms the token-local branch
$B_{u,0}=Y_{u,1}.$
The default configuration with two learnable windows (\emph{2LW}) adds two Gaussian-mixture branches. For branch $s\in\{1,2\}$, the receptive-field ratio is
$r_{u,s}=\operatorname{clip}\left(\rho_{u,s}+0.5\tanh\left(\theta_{u,s}\right),10^{-3},1-10^{-3}\right).$
The premise anchor ratios are
$(\rho_{p,1},\rho_{p,2})=(0.17,0.33),$
whereas the hypothesis anchor ratios are
$(\rho_{h,1},\rho_{h,2})=(0.30,0.65).$
These anchors initialize the receptive-field ratios, while the offset parameters are initialized as
$
\theta_{u,s}
\sim
\mathcal{N}
\left(
0,
0.5^{2}
\right),
$
and transformed by the bounded $\tanh$ term to determine the trainable effective ratios.

For a minibatch of $N_{\mathrm{batch}}$ examples, define the mean active length as
$n_{u}^{\mathrm{batch}}=\frac{1}{N_{\mathrm{batch}}}\sum_{i=1}^{N_{\mathrm{batch}}}n_{u}^{(i)}.$
Each module initializes its running statistic with the first minibatch and subsequently updates it as
$
\overline{n}_{u}
\leftarrow
0.95\overline{n}_{u}
+
0.05n_{u}^{\mathrm{batch}}.
$
The statistic is updated during training and fixed during evaluation.
We set the target kernel center at
$
\mu_{u,s}
=
\operatorname{clip}
\left(
r_{u,s}\overline{n}_{u},
\min\mathcal{K}_{u},
\max\mathcal{K}_{u}-0.5
\right).
$
We fix the Gaussian bandwidth at $\sigma=5$. The first mixture excludes the isolated $\kappa=1$ candidate,
$
\mathcal{K}_{u,1}
=
\mathcal{K}_{u}
\setminus
\{1\},
$
while the second uses the full bank,
$
\mathcal{K}_{u,2}
=
\mathcal{K}_{u}.
$
For $s\in\{1,2\}$ and $\kappa\in\mathcal{K}_{u,s}$, the normalized weight is
$\gamma_{u,s,\kappa}=\frac{\exp\left(-\frac{(\kappa-\mu_{u,s})^{2}}{2\sigma^{2}}\right)}{\sum_{\kappa'\in\mathcal{K}_{u,s}}\exp\left(-\frac{(\kappa'-\mu_{u,s})^{2}}{2\sigma^{2}}\right)}.$
The resulting branch feature is
$B_{u,s}=\sum_{\kappa\in\mathcal{K}_{u,s}}\gamma_{u,s,\kappa}Y_{u,\kappa}.$
The global branch mean-pools the segment tensor and applies a learned linear projection:
$g_{u}=\frac{1}{n_u^{\mathrm{pad}}}\sum_{j=1}^{n_u^{\mathrm{pad}}}\widetilde{E}_{j}^{u},$
$B_{u,3,j}=W_{u,\mathrm{global}}g_{u}+b_{u,\mathrm{global}}.$
Each branch is then mapped independently to a CVTR as
$Z_{u,s}=\Phi_{\mathrm{CVTR}}\left(B_{u,s}\right).$

Let $c_{u,s}$ be the segment-level fusion logits. The corresponding weights are
$
\omega_{u,s}
=
\frac{
	\exp
	\left(
	c_{u,s}
	\right)
}{
	\sum_{v=0}^{3}
	\exp
	\left(
	c_{u,v}
	\right)
}.
$
We define the shared reference as
$R^{u}=\sum_{s=0}^{3}\omega_{u,s}Z_{u,s}+P^{u},$
where ${P}^{u}$ is a real-valued sinusoidal positional encoding, zeroed at padded positions. Zero-initialized fusion logits assign equal initial weights to the four branches.
We concatenate the premise and hypothesis references as
$
{Z}^{(0)}
=
\left[
{R}^{p};
{R}^{h}
\right].
$ 
Kernel grids, anchors, and bandwidth remain fixed, while effective ratios, Gaussian-mixture weights, and fusion weights are learned.

\subsection{Layer-Specific Conditioning}
\label{sec:tct_residual}

Each layer applies separate multiscale convolutional modules to the real and imaginary reference components. All modules use the candidate kernel set
$
\mathcal{K}
=
\{1,3,7,13,21,33\},
$
anchor initialization $(0.17,0.33)$, and fixed bandwidth $\sigma=5$. Each module contains an isolated $\kappa=1$ branch, two Gaussian-mixture branches, and one global branch.
The candidate kernel sizes are fixed across layers, while the convolutional filters, receptive-field offsets, running length statistics, and branch-fusion logits are learned separately for each layer and component. Because these modules process the concatenated premise--hypothesis reference, both sequences share the same candidate kernel bank.
Let $B_{\tau}^{(\ell,s)}$ denote branch $s$ produced by the layer-$\ell$ module for component $\tau\in\{\mathrm{real},\mathrm{imag}\}$. The layer-specific branch weights are
$\beta_{\tau}^{(\ell,s)}=\frac{\exp\left(c_{\tau}^{(\ell,s)}\right)}{\sum_{v=0}^{3}\exp\left(c_{\tau}^{(\ell,v)}\right)}.$
The logits $c_{\tau}^{(\ell,s)}$ are initialized to zero, giving a uniform initial mixture.
The fused conditioning feature is
$\widehat{B}_{\tau}^{(\ell)}=\sum_{s=0}^{3}\beta_{\tau}^{(\ell,s)}B_{\tau}^{(\ell,s)}.$

Given the layer input $Z^{(\ell)}$, the conditioned components are
$X_{\mathrm{real}}^{(\ell)}=\operatorname{Re}\left(Z^{(\ell)}\right)+\xi_{\mathrm{real}}^{(\ell)}\widehat{B}_{\mathrm{real}}^{(\ell)},$
$X_{\mathrm{imag}}^{(\ell)}=\operatorname{Im}\left(Z^{(\ell)}\right)+\xi_{\mathrm{imag}}^{(\ell)}\widehat{B}_{\mathrm{imag}}^{(\ell)}.$
Each $\xi_{\tau}^{(\ell)}$ is a zero-initialized learned scalar, so the conditioning contribution begins at zero.

\subsection{Token--Context Attention under SMIC Conditioning}
\label{sec:tct_update}
We project queries and values from the normalized real component, and project keys from the imaginary component:
$
\overline{{X}}_{\mathrm{real}}^{(\ell)}
=
\operatorname{LN}_{1}^{(\ell)}
\left(
{X}_{\mathrm{real}}^{(\ell)}
\right),
$
$
{Q}_{a}^{(\ell)}
=
\overline{{X}}_{\mathrm{real}}^{(\ell)}
{W}_{Q,a}^{(\ell)},
$
$
{K}_{a}^{(\ell)}
=
{X}_{\mathrm{imag}}^{(\ell)}
{W}_{K,a}^{(\ell)},
$
$
{V}_{a}^{(\ell)}
=
\overline{{X}}_{\mathrm{real}}^{(\ell)}
{W}_{V,a}^{(\ell)}.
$
Let ${M}_{a}^{(\ell)}$ denote the causal and key-padding mask for head $a$. The attention distribution is
$
{A}_{a}^{(\ell)}
=
\text{Softmax}
\left(
\frac{
	{Q}_{a}^{(\ell)}
	\left(
	{K}_{a}^{(\ell)}
	\right)^{\top}
}{
	\sqrt{d_a}
}
+
{M}_{a}^{(\ell)}
\right).
$
Rows corresponding to padded queries are zeroed. The multi-head output is then
$
{O}^{(\ell)}
=
\operatorname{Concat}_{a=1}^{H}
\left(
\operatorname{Dropout}
\left(
{A}_{a}^{(\ell)}
\right)
{V}_{a}^{(\ell)}
\right)
{W}_{O}^{(\ell)}.
$
After computing the multi-head output, the residual branch updates the real component as
$
{U}^{(\ell)}
=
{X}_{\mathrm{real}}^{(\ell)}
+
\operatorname{Dropout}
\left(
\operatorname{LN}_{2}^{(\ell)}
\left(
{O}^{(\ell)}
\right)
\right).
$
The feed-forward sublayer then processes the resulting representation:
\[
{F}^{(\ell)}
=
\operatorname{Dropout}
\left(
\operatorname{ReLU}
\left(
{U}^{(\ell)}
{W}_{1}^{(\ell)}
+
{b}_{1}^{(\ell)}
\right)
{W}_{2}^{(\ell)}
+
{b}_{2}^{(\ell)}
\right).
\]
Adding this feed-forward contribution yields the next-layer real component,
$
{T}^{(\ell+1)}
=
{U}^{(\ell)}
+
{F}^{(\ell)}.
$
In contrast, the conditioned imaginary component is propagated without further transformation:
$
{C}^{(\ell+1)}
=
{X}_{\mathrm{imag}}^{(\ell)}.
$
The two components are then recombined to form the next complex representation:
$Z^{(\ell+1)}=T^{(\ell+1)}+\mathrm{i}C^{(\ell+1)}.$
TCT-SMIC has three attention layers, $32$ heads, and a feed-forward width of $4096$. Classification uses the final-layer real component at the last valid hypothesis position.

\begin{table*}[!t]
	\centering
	\caption{ANLI accuracy and training FLOPs across parameter-comparable configurations.}
	\label{exp2}
	\footnotesize
	\setlength{\tabcolsep}{2pt}
	\renewcommand{\arraystretch}{0.95}
	
	\begin{tabular}{
			@{}
			>{\raggedright\arraybackslash}m{2.8cm}
			*{3}{
				>{\centering\arraybackslash}m{1.8cm}
				>{\centering\arraybackslash}m{1.6cm}
			}
			@{}
		}
		\toprule
		\textbf{Model}
		& \shortstack{\textbf{R1}\\\textbf{Acc/FLOPs}}
		& \shortstack{\textbf{Diff}\\\textbf{[95\% CI]}}
		& \shortstack{\textbf{R2}\\\textbf{Acc/FLOPs}}
		& \shortstack{\textbf{Diff}\\\textbf{[95\% CI]}}
		& \shortstack{\textbf{R3}\\\textbf{Acc/FLOPs}}
		& \shortstack{\textbf{Diff}\\\textbf{[95\% CI]}} \\
		\midrule
		
		\multicolumn{7}{@{}l}{\textbf{0.8B configurations}} \\
		\midrule
		
		\textbf{TCT-SMIC (0.80B)}
		& \textbf{39.03}/1.06k & --
		& \textbf{44.89}/2.54k & --
		& \textbf{43.17}/4.27k & -- \\
		
		Phi-1.5-style (0.82B)
		& 36.88/1.37k
		& \shortstack{2.15\\{}[0.88,3.42]}
		& 41.13/2.60k
		& \shortstack{3.76\\{}[2.67,4.85]}
		& 41.59/8.04k
		& \shortstack{1.58\\{}[0.83,2.32]} \\
		
		Llama-3-style (0.85B)
		& 36.61/1.22k
		& \shortstack{2.42\\{}[0.86,3.98]}
		& 41.12/2.30k
		& \shortstack{3.77\\{}[2.52,5.02]}
		& 41.23/5.11k
		& \shortstack{1.94\\{}[0.92,2.96]} \\
		
		Pythia-style (0.81B)
		& 36.02/1.23k
		& \shortstack{3.01\\{}[1.95,4.07]}
		& 38.17/3.00k
		& \shortstack{6.72\\{}[5.16,8.28]}
		& 40.44/9.77k
		& \shortstack{2.73\\{}[1.33,4.12]} \\
		
		MLP (0.80B)
		& \emph{37.57}/0.99k
		& \shortstack{1.46\\{}[0.25,2.67]}
		& \emph{43.43}/3.42k
		& \shortstack{1.46\\{}[0.42,2.50]}
		& \textit{41.83}/6.18k
		& \shortstack{1.34\\{}[0.29,2.40]} \\
		
		\midrule
		\multicolumn{7}{@{}l}{\textbf{1B configurations}} \\
		\midrule

		Phi-1.5-style (1.3B)
		& \emph{36.76}/2.21k
		& \shortstack{2.27\\{}[0.94,3.60]}
		& 40.75/4.34k
		& \shortstack{4.14\\{}[3.12,5.16]}
		& 41.72/13.5k
		& \shortstack{1.45\\{}[0.12,2.76]} \\
		
		Llama-3-style (1.23B)
		& 36.52/1.95k
		& \shortstack{2.51\\{}[1.78,3.24]}
		& 41.28/3.79k
		& \shortstack{3.61\\{}[2.10,5.17]}
		& 40.28/7.80k
		& \shortstack{2.88\\{}[2.08,3.69]} \\
		
		Pythia-style (1.08B)
		& 35.69/1.99k
		& \shortstack{3.34\\{}[1.81,4.87]}
		& 39.55/5.58k
		& \shortstack{5.34\\{}[3.13,7.55]}
		& 39.08/10.9k
		& \shortstack{4.09\\{}[1.79,6.38]} \\
		
		Gemma-3-style (1.0B)
		& 35.65/1.11k
		& \shortstack{3.38\\{}[1.47,5.29]}
		& \emph{42.44}/2.58k
		& \shortstack{2.45\\{}[1.08,3.82]}
		& \emph{42.39}/4.83k
		& \shortstack{0.78\\{}[0.03,1.52]} \\
		
		\bottomrule
	\end{tabular}
	
	\par\smallskip
	\begin{minipage}{\linewidth}
		\footnotesize
		Each Acc/FLOPs entry reports accuracy (\%) followed by estimated
		training FLOPs (PF; k denotes $10^3$).
		Differences and confidence intervals are in percentage points.
		Boldface denotes TCT-SMIC, and italics denote the strongest baseline within each parameter group and ANLI round.
	\end{minipage}
\end{table*}

\section{Results} \label{sec:exp}

  All ANLI accuracy comparisons use models trained from random initialization. For the parameter-comparable configurations in Table~\ref{exp2}, we use FP32 arithmetic and deterministic CUDA and cuDNN kernels, with TF32 and cuDNN benchmarking disabled. Each configuration uses its default tokenizer. We tune learning rate and dropout on ANLI-R1 using five seeds and select the configuration with the highest mean development accuracy.   Appendix~\ref{app:settings} lists the optimizer, batch size, search ranges, selected values, and the FLOP estimation procedure. The selected hyperparameters are then reused for ANLI-R2 and ANLI-R3. Each ANLI round is trained and early-stopped separately with a patience of $5$. We evaluate five additional seeds and report ten-seed mean accuracies with seed-paired $95\%$ confidence intervals for pairwise differences.   The from-scratch analyses and ablations use the same seeds. We evaluate pretrained ModernBERT and Llama backbones with and without SMIC through deletion/insertion diagnostics. All reported runs use NVIDIA GH200 accelerators with eight CPU cores and 128 GB of host memory.

\subsection{Performance comparisons on ANLI}\label{sec:anliresult}

Table~\ref{exp2} reports test accuracy on ANLI-R1, ANLI-R2, and ANLI-R3. TCT-SMIC achieves the highest mean accuracy among all configurations in Table~\ref{exp2} on each ANLI round. It also has lower estimated training FLOPs than most baselines in each round.

\subsection{Attention-Based Diagnostics under Fixed-Length Token Perturbations}
\label{sec:attnfaith}

We use fixed-length token-ID replacements for the attention diagnostics.
We sample $1000$ labeled ANLI-R1 development examples with seed $42$. For TCT-SMIC, we average results across checkpoints trained with the same seeds. Premises and hypotheses are tokenized separately with the Llama-3.2-1B tokenizer, with maximum lengths of $110$ and $40$, respectively.
We replace token IDs after tokenization while preserving sequence length, attention masks, and token indices. The original and perturbed attention distributions therefore share the same token-index set.

For each replacement count $n_{\mathrm{rep}}\in\{1,2,4\}$, we generate three deterministic realizations per example and condition. The stopword-targeted condition samples eligible single-token stopwords, falling back to all eligible non-special positions when none are available. Replacement IDs are drawn from non-special entries that decode to alphabetic strings of at least two characters. If fewer than $n_{\mathrm{rep}}$ positions are available, all available positions are replaced. The random condition samples positions from all eligible non-special positions.

\paragraph{Deletion and insertion.}

Deletion and insertion are evaluated on the unmodified inputs. For each plotted configuration, we average final-layer attention across heads. 
Let $\mathbf{A}_i$ be the head-averaged attention matrix for example $i$, and let $L_i$ denote the query position used to extract the attention ranking for the evaluated configuration. For TCT-SMIC, $L_i$ is the final valid hypothesis position. Candidate positions $j$ are ranked by their attention weight $A_{i,L_i,j}$.
Let $\mathcal{D}_i$ be the set of non-special positions excluding $L_i$. We define
$n_i^{\mathrm{top}}=\left\lceil 0.1\left|\mathcal{D}_i\right|\right\rceil.$
Let $\operatorname{conf}_{i,0}$ be the confidence of the original prediction. Replacing the top $t$ ranked positions with a fixed non-special placeholder yields a normalized confidence drop
$\psi_{i,t}^{\mathrm{del}}=\frac{\max\left\{0,\operatorname{conf}_{i,0}-\operatorname{conf}_{i,t}^{\mathrm{del}}\right\}}{\operatorname{conf}_{i,0}}.$

For insertion, all positions in $\mathcal{D}_i$ are first replaced by the placeholder, after which the top $t$ ranked positions are restored:
$\psi_{i,t}^{\mathrm{ins}}=\operatorname{clip}\left(\frac{\operatorname{conf}_{i,t}^{\mathrm{ins}}}{\operatorname{conf}_{i,0}},0,1\right).$
Let
$0=t_{i,0}<t_{i,1}<\cdots<t_{i,\Lambda_i}=n_i^{\mathrm{top}}$
denote the evaluated intervention steps, and let
$f_{i,\lambda}=\frac{t_{i,\lambda}}{n_i^{\mathrm{top}}}.$
The deletion and insertion AUCs are
$\operatorname{AUC}_{i,\mathrm{del}}=\sum_{\lambda=1}^{\Lambda_i}\left(f_{i,\lambda}-f_{i,\lambda-1}\right)\frac{\psi^{\mathrm{del}}_{i,t_{i,\lambda-1}}+\psi^{\mathrm{del}}_{i,t_{i,\lambda}}}{2}$
and
$\operatorname{AUC}_{i,\mathrm{ins}}=\sum_{\lambda=1}^{\Lambda_i}\left(f_{i,\lambda}-f_{i,\lambda-1}\right)\frac{\psi^{\mathrm{ins}}_{i,t_{i,\lambda-1}}+\psi^{\mathrm{ins}}_{i,t_{i,\lambda}}}{2}.$
The combined per-example diagnostic is
$
\operatorname{DI\text{-}AUC}_{i,0.1}
=
\frac{1}{2}
\left(
\operatorname{AUC}_{i,\mathrm{del}}
+
\operatorname{AUC}_{i,\mathrm{ins}}
\right).
$
The reported score is
$
\operatorname{DI\text{-}AUC}_{0.1}
=
\frac{1}{N}
\sum_{i=1}^{N}
\operatorname{DI\text{-}AUC}_{i,0.1}.
$

\paragraph{Conditional pooled-attention JSD.}

For each example $i$, let $\Omega_i=\mathcal{D}_i$ be the common token-index set excluding special tokens and the final pooling token. Let $\tilde{p}_{i,j}$ and $\tilde{q}_{i,j}$ denote the head-averaged original and perturbed pooled-query attention weights normalized over valid positions. The metric is defined when
$\left|\Omega_i\right|\ge 2.$
Restricting the weights to $\Omega_i$, clipping them below at $\epsilon=10^{-8}$, and renormalizing yields distributions $\mathbf{p}_i$ and $\mathbf{q}_i$.
For every $j\in\Omega_i$,
$p_{i,j}=\frac{\max\left(\tilde{p}_{i,j},\epsilon\right)}{\sum_{v\in\Omega_i}\max\left(\tilde{p}_{i,v},\epsilon\right)},$
$q_{i,j}=\frac{\max\left(\tilde{q}_{i,j},\epsilon\right)}{\sum_{v\in\Omega_i}\max\left(\tilde{q}_{i,v},\epsilon\right)}.$
Using natural logarithms, define
$
\mathbf{m}_i
=
\frac{
	\mathbf{p}_i
	+
	\mathbf{q}_i
}{
	2
},
$
$
J_i
=
\frac{1}{2}
\operatorname{KL}
\left(
\mathbf{p}_i
\mathbin{\|}
\mathbf{m}_i
\right)
+
\frac{1}{2}
\operatorname{KL}
\left(
\mathbf{q}_i
\mathbin{\|}
\mathbf{m}_i
\right).
$
We report
$
\overline{J}
=
\frac{1}{N}
\sum_{i=1}^{N}
J_i.
$

\begin{figure}[t]
	\centering
	\captionsetup[subfigure]{skip=2pt}
	
	\begin{subfigure}[t]{0.249\linewidth}
		\centering
		\includegraphics[width=\linewidth]{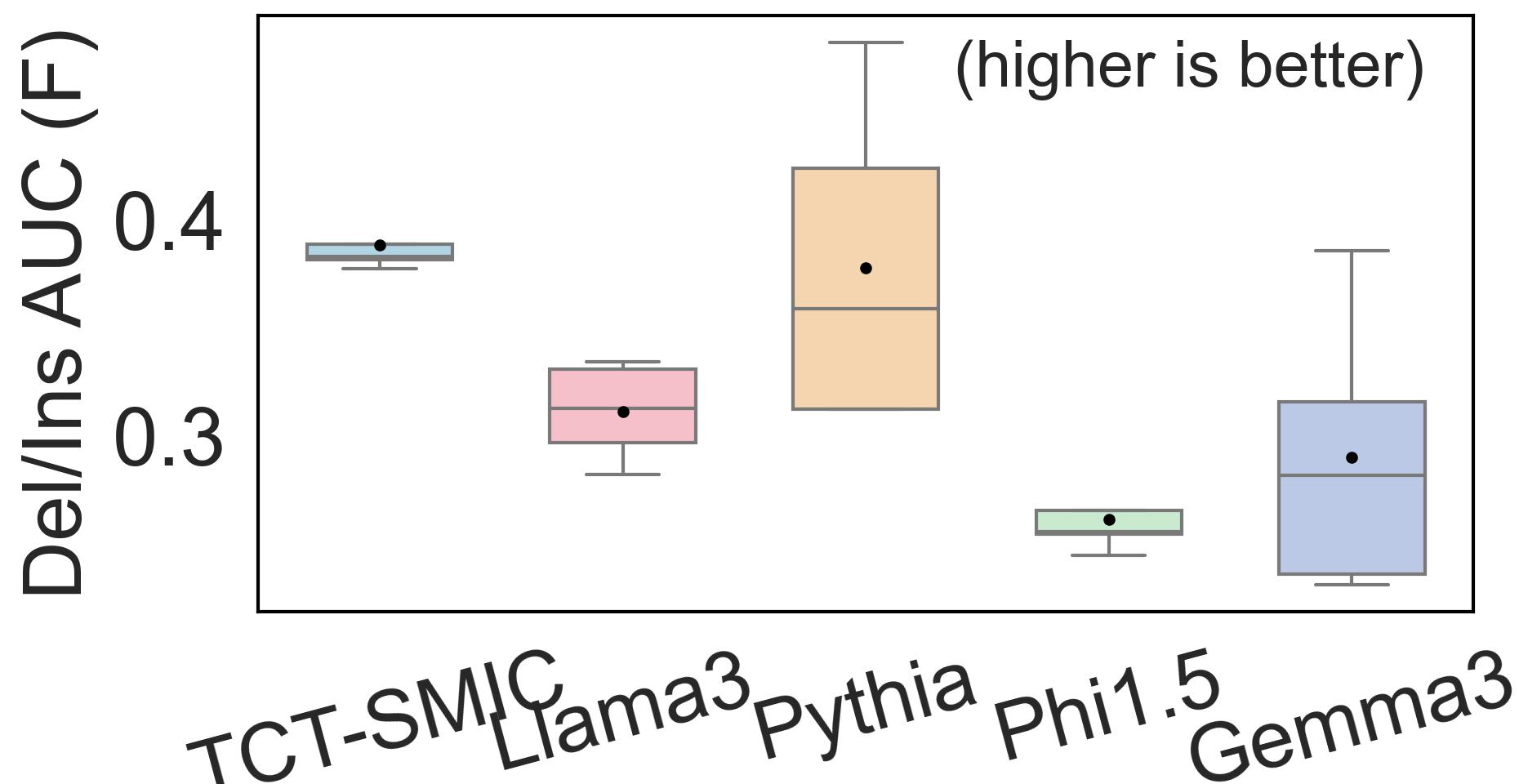}
		\caption{Baseline AUC.}
		\label{fig:attn_baseline_auc}
	\end{subfigure}\hfill
	\begin{subfigure}[t]{0.249\linewidth}
		\centering
		\includegraphics[width=\linewidth]{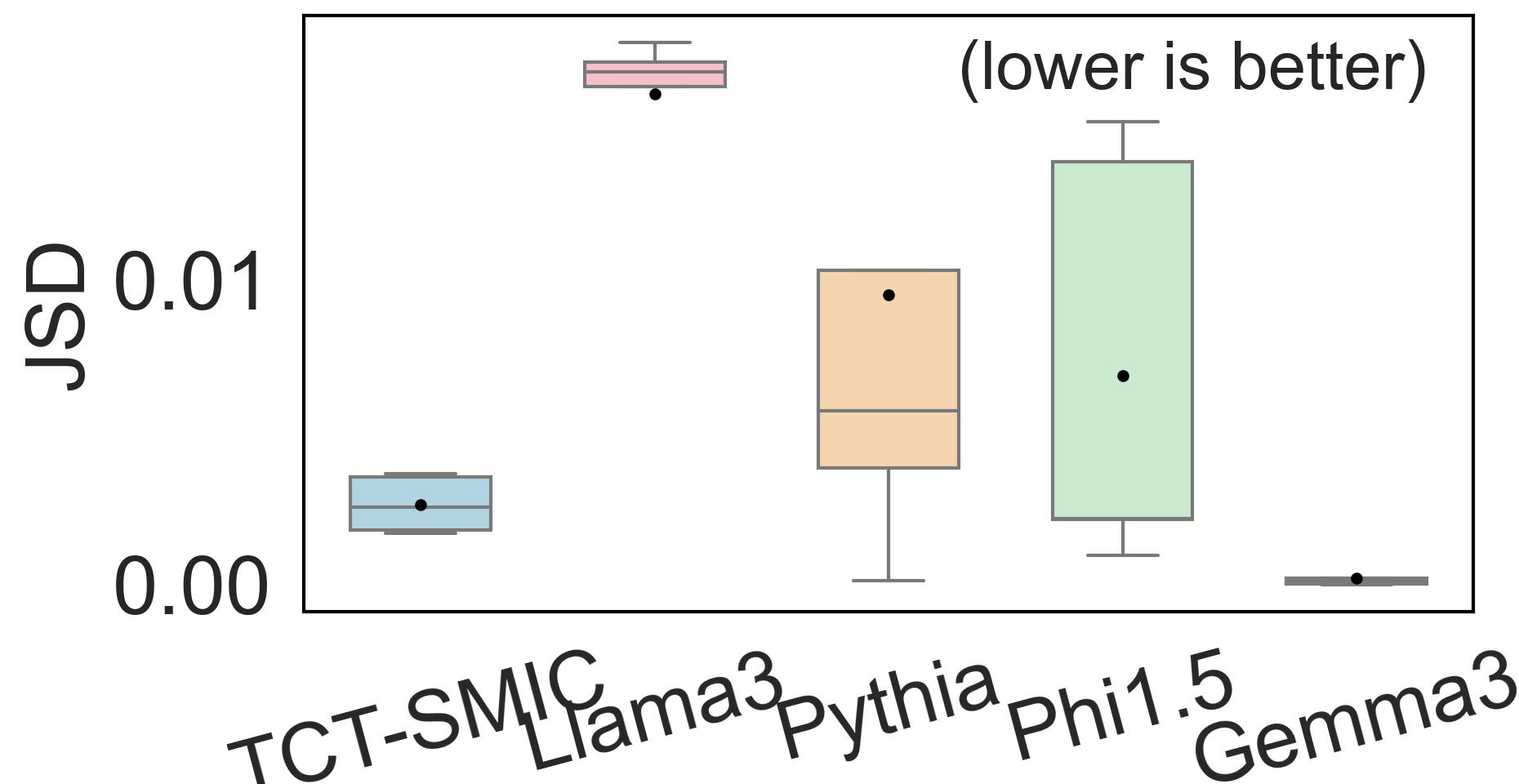}
		\caption{Baseline JSD.}
		\label{fig:attn_baseline_jsd}
	\end{subfigure}\hfill
	\begin{subfigure}[t]{0.249\linewidth}
		\centering
		\includegraphics[width=\linewidth]{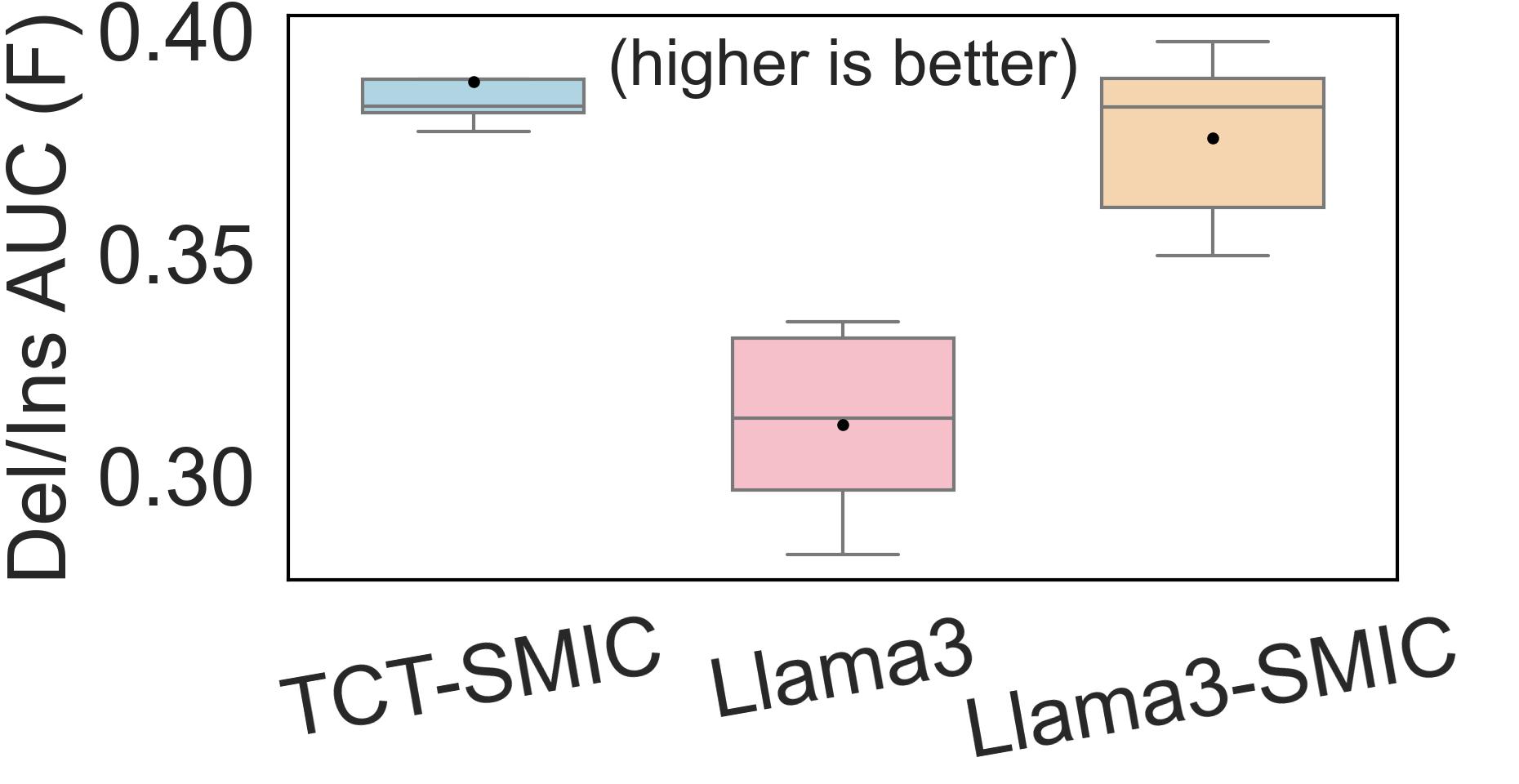}
		\caption{Llama-3-style AUC.}
		\label{fig:attn_llama_auc}
	\end{subfigure}\hfill
	\begin{subfigure}[t]{0.249\linewidth}
		\centering
		\includegraphics[width=\linewidth]{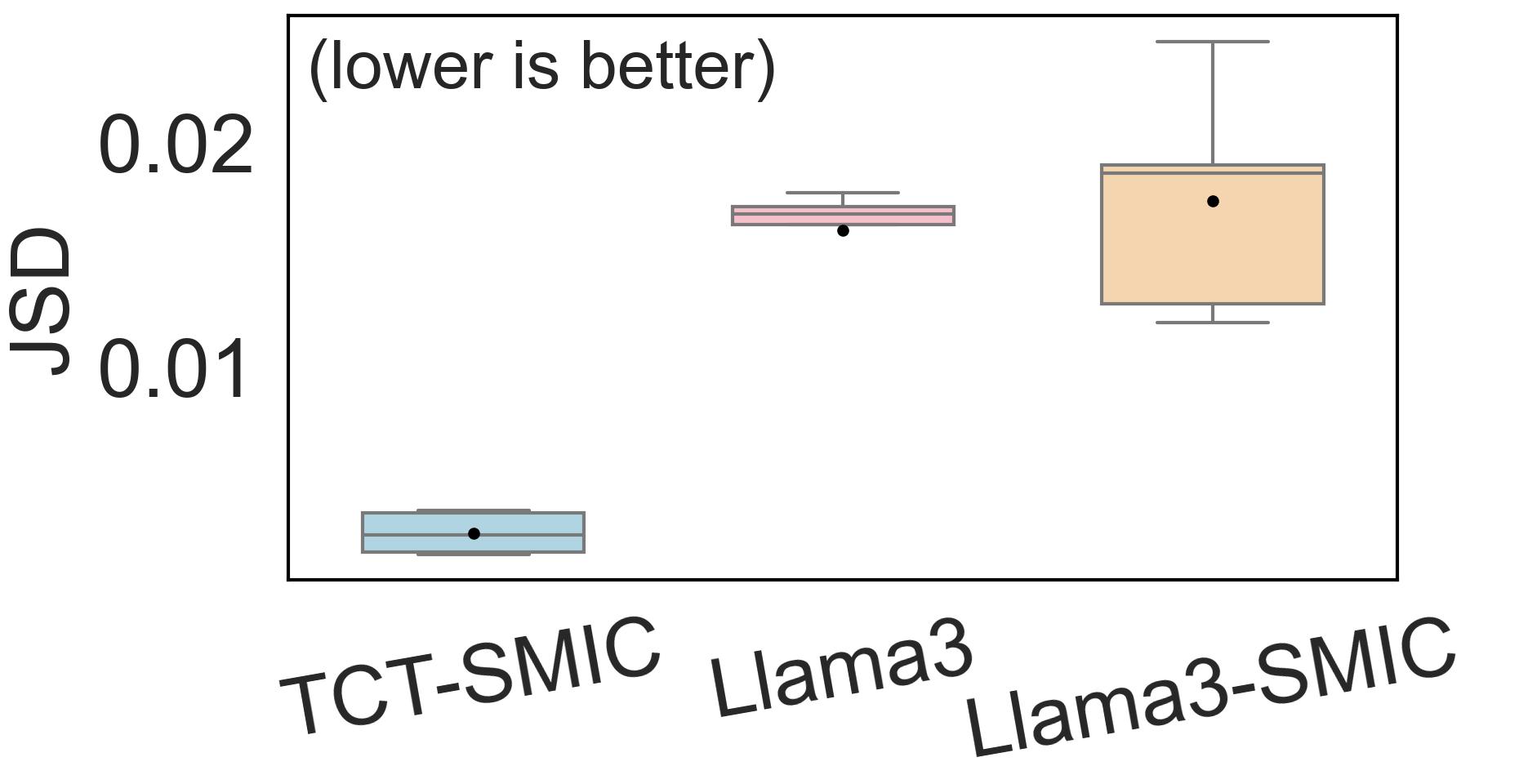}
		\caption{Llama-3-style JSD.}
		\label{fig:attn_llama_jsd}
	\end{subfigure}
	
	\caption{Attention diagnostics on ANLI-R1.}
	\label{fig:attn_all}
\end{figure}
\begin{proposition}[Conditional Subset-Mass Bound]
	\label{prop:js_subset_mass}
	
	For any collection of subsets $\mathcal{S}_i\subseteq\Omega_i$,
	$
	\frac{1}{N}
	\sum_{i=1}^{N}
	\left|
	p_i
	\left(
	\mathcal{S}_i
	\right)
	-
	q_i
	\left(
	\mathcal{S}_i
	\right)
	\right|
	\le
	\min
	\left\{
	1,
	\sqrt{
		2\overline{J}
	}
	\right\}.
	$
\end{proposition}
The proof is provided in Appendix~\ref{prop:js_subset_mass1}.
Proposition~\ref{prop:js_subset_mass} applies directly to the top-$10\%$ prefix $\mathcal{S}_{i}^{0.1}$ formed by the $n_i^{\mathrm{top}}$ highest-ranked positions under $\mathbf{p}_i$. The bound holds for any model, so the conditional JSD values of all configurations in Figure~\ref{fig:attn_baseline_jsd} can be read on the same scale.

Deletion/insertion AUC measures reliance on the most-attended tokens, whereas conditional JSD measures attention redistribution after an edit. Figures~\ref{fig:attn_baseline_auc} and~\ref{fig:attn_baseline_jsd} show that TCT-SMIC combines relatively high top-$10\%$ deletion/insertion AUC with low conditional JSD.

\paragraph{SMIC as a plug-in for the Llama-3-style configuration.}
\label{llwithwin}
We add a real-valued SMIC module to the Llama-3-style configuration to test its applicability beyond complex-valued TCT-SMIC. All other components and training settings are unchanged. Figure~\ref{fig:attn_llama_auc} shows that adding SMIC increases the deletion/insertion summary score, while Figure~\ref{fig:attn_llama_jsd} reports the corresponding conditional pooled-attention JSD.

\paragraph{SMIC in pretrained backbones.}

\begin{figure}[t]
	\centering
	\begin{subfigure}[t]{0.48\linewidth}
		\centering
		\includegraphics[width=0.9\linewidth]{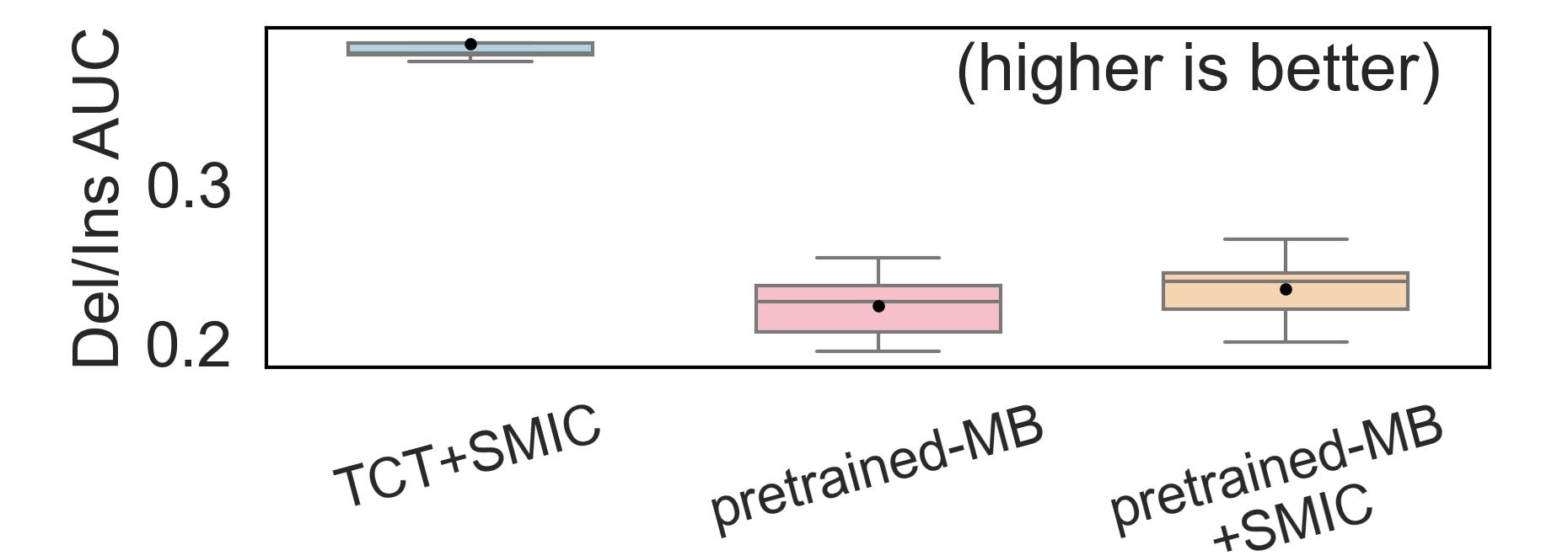}
		\caption{Pretrained ModernBERT.}
		\label{fig:pretrained_attn}
	\end{subfigure}
	\hfill
	\begin{subfigure}[t]{0.49\linewidth}
		\centering
		\includegraphics[width=0.9\linewidth]{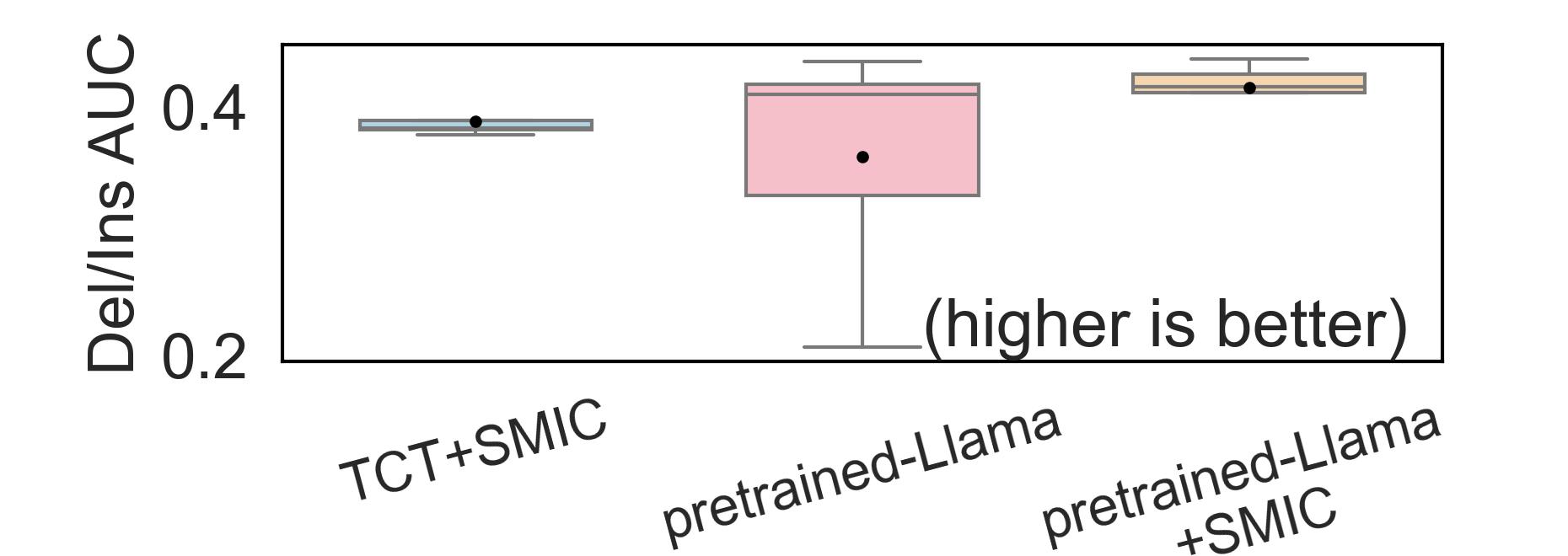}
		\caption{Pretrained Llama.}
		\label{fig:pretrained_llama}
	\end{subfigure}
	\caption{Top-$10\%$ deletion/insertion AUC on ANLI-R1
		for pretrained models with and w/o SMIC.
	}
	\label{fig:pretrained_smic}
\end{figure}

We evaluate pretrained ModernBERT and Llama backbones with and without SMIC using the top-$10\%$ deletion/insertion diagnostic. Figures~\ref{fig:pretrained_attn} and~\ref{fig:pretrained_llama} show a higher median score with SMIC for both backbones. Llama also exhibits a narrower interquartile range. These results show that the observed diagnostic effect is present in both pretrained encoder and decoder backbones and is not restricted to training from random initialization.

\begin{figure*}[htbp]
	\center
	\includegraphics[width=0.99\linewidth,height=2.6cm]{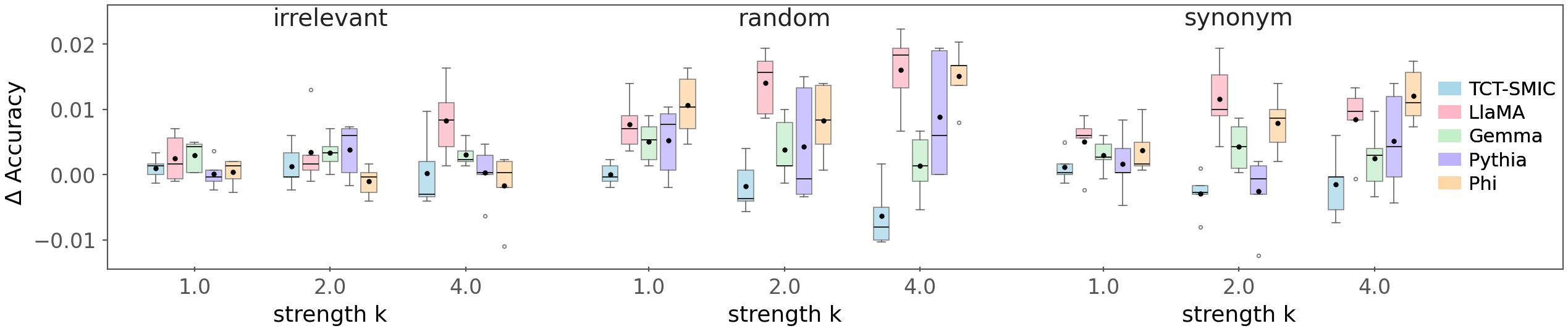}
	\caption{Signed accuracy change on ANLI-R1 under surface-text perturbations. The horizontal axis shows the replacement count $n_{\mathrm{rep}}$.}
	\label{fig:accdrop}
\end{figure*}

\subsection{Empirical sensitivity under sparse lexical perturbations}\label{sec:work-pertur}

The signed accuracy analysis uses pre-generated surface-text perturbations. This protocol differs from the fixed-length token-ID protocol used for the attention diagnostics in Section~\ref{sec:attnfaith}.

After removing unlabeled examples, we sample $1000$ labeled examples from the ANLI-R1 development set with seed 42. The pre-generated data contain three perturbation families and replacement counts $n_{\mathrm{rep}}\in\{1,2,4\}$.
Each value of $n_{\mathrm{rep}}$ has three perturbation realizations.

The \textsc{Synonym} family uses WordNet near-synonyms as a relatively conservative intervention. The data label function-word and punctuation replacements as \textsc{Irrelevant}. The \textsc{Random} family uses unconstrained lexical replacements and is treated as a corruption stress test.

The perturbations are generated once at the surface-text level, and the same perturbed premise--hypothesis strings are evaluated across models. Each configuration tokenizes the original and perturbed strings with its own tokenizer. 

For every model and condition, the reported signed accuracy change is
$
\Delta\operatorname{Acc}
=
\operatorname{Acc}_{\mathrm{original}}
-
\operatorname{Acc}_{\mathrm{perturbed}}.
$
Positive values indicate lower accuracy after perturbation, whereas negative values indicate higher accuracy after perturbation.

\paragraph{Finite-sample interpretation of signed accuracy change.}

Fix one model configuration and one perturbation condition. Consider $N$ evaluated pairs $(x_i,x_i')$, where $x_i$ is the original input, $x_i'$ is its perturbed counterpart, and $y_i$ is the retained evaluation label. Let $\widehat{y}_i$ and $\widehat{y}_i'$ denote the corresponding predictions. Then
$\Delta\operatorname{Acc}=\frac{1}{N}\sum_{i=1}^{N}\left(\mathbf{1}\{\widehat{y}_i=y_i\}-\mathbf{1}\{\widehat{y}_i'=y_i\}\right).$

Let $\mathcal{B}=\{i:\widehat{y}_i=y_i,\ \widehat{y}_i'\ne y_i\}$ denote predictions that become incorrect after perturbation, and let $\mathcal{G}=\{i:\widehat{y}_i\ne y_i,\ \widehat{y}_i'=y_i\}$ denote predictions that become correct. Then the signed accuracy change satisfies
$\Delta\operatorname{Acc}=\frac{|\mathcal{B}|-|\mathcal{G}|}{N}.$

Let $\mathcal{W}=\{i:\widehat{y}_i\ne\widehat{y}_i',\ \widehat{y}_i\ne y_i,\ \widehat{y}_i'\ne y_i\}$ denote wrong-to-wrong prediction transitions. The prediction-flip rate is
$
\operatorname{FlipRate}
=
\frac{
	|\mathcal{B}|
	+
	|\mathcal{G}|
	+
	|\mathcal{W}|
}{
	N
}. 
$
The signed accuracy change can therefore be written as
$
\Delta\operatorname{Acc}
=
\operatorname{BreakRate}
-
\operatorname{RecoveryRate},
$
where
$
\operatorname{BreakRate}
=
\frac{
	|\mathcal{B}|
}{
	N
}
$
and
$
\operatorname{RecoveryRate}
=
\frac{
	|\mathcal{G}|
}{
	N
}.
$
Thus, $\Delta\operatorname{Acc}$ is the break rate minus the recovery rate, whereas $\operatorname{FlipRate}$ also includes wrong-to-wrong transitions.

Signed accuracy change is not the proportion of changed predictions. 
Because the perturbed examples are not manually re-annotated, $\Delta\operatorname{Acc}$ can reflect both model sensitivity and changes in the correct NLI relation. 

  Figure~\ref{fig:accdrop} shows that TCT-SMIC has the smallest mean signed accuracy change across the ANLI-R1 conditions. For the synonym and function-word or punctuation families, this means a smaller net accuracy change. For random replacements, it means a smaller accuracy drop under unconstrained corruption.

\begin{figure}
	\centering
	\includegraphics[
	width=1\linewidth,
	height=3.1cm,
	keepaspectratio
	]{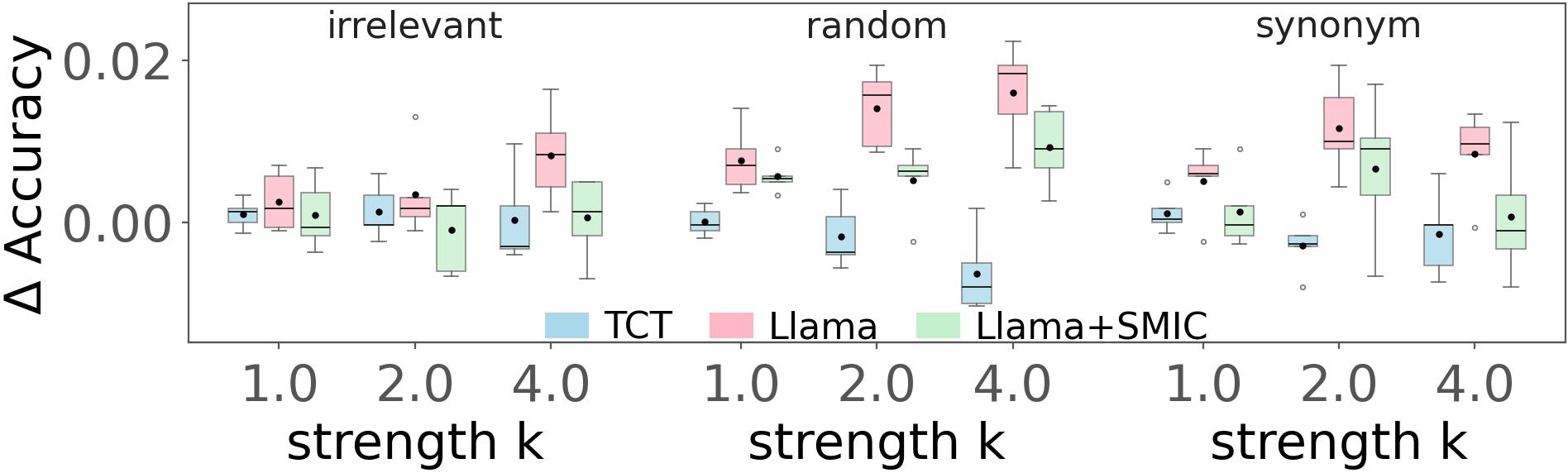}
	\caption{Signed accuracy change for TCT-SMIC (TCT in the legend) and the Llama-3-style configurations with and without SMIC on the evaluated ANLI-R1 surface-text perturbations. The horizontal axis shows the replacement count $n_{\mathrm{rep}}$.}
	\label{fig:llwin}
\end{figure}

\paragraph{SMIC as a plug-in for the Llama-3-style configuration.}
Figure~\ref{fig:llwin} shows that adding SMIC to the Llama-3-style configuration reduces the median $\Delta\mathrm{Acc}$ across the evaluated substitution families.

\subsection{Ablation}\label{sec:abla}

We test each part of SMIC by removing it or replacing it with a simpler design. Removing SMIC tests the module as a whole. A token-level reference, computed directly from token embeddings without multiscale features, tests the multiscale construction and is the closest variant to plain input injection. Varying the number of learnable scales from zero to three tests the learned scale mixture, and the zero-scale variant keeps only the token-local and global branches. A layer-local variant tests whether the reference must come from the input and be shared across layers. The real-valued Llama-style plug-in in Sections~\ref{sec:attnfaith} and~\ref{sec:work-pertur} tests SMIC without CVTR and TCA. We also report depth-matched and tokenizer checks.

\begin{figure}[t]
	\centering
	\includegraphics[width=0.49\linewidth]{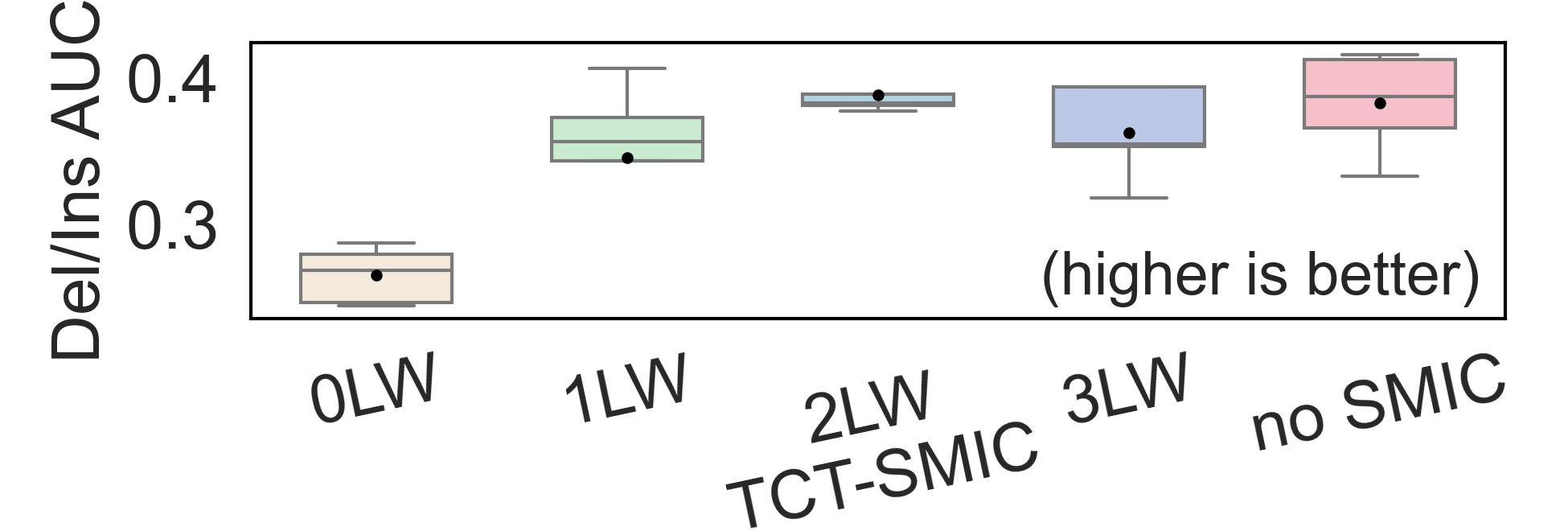}
	\hfill
	\includegraphics[width=0.49\linewidth]{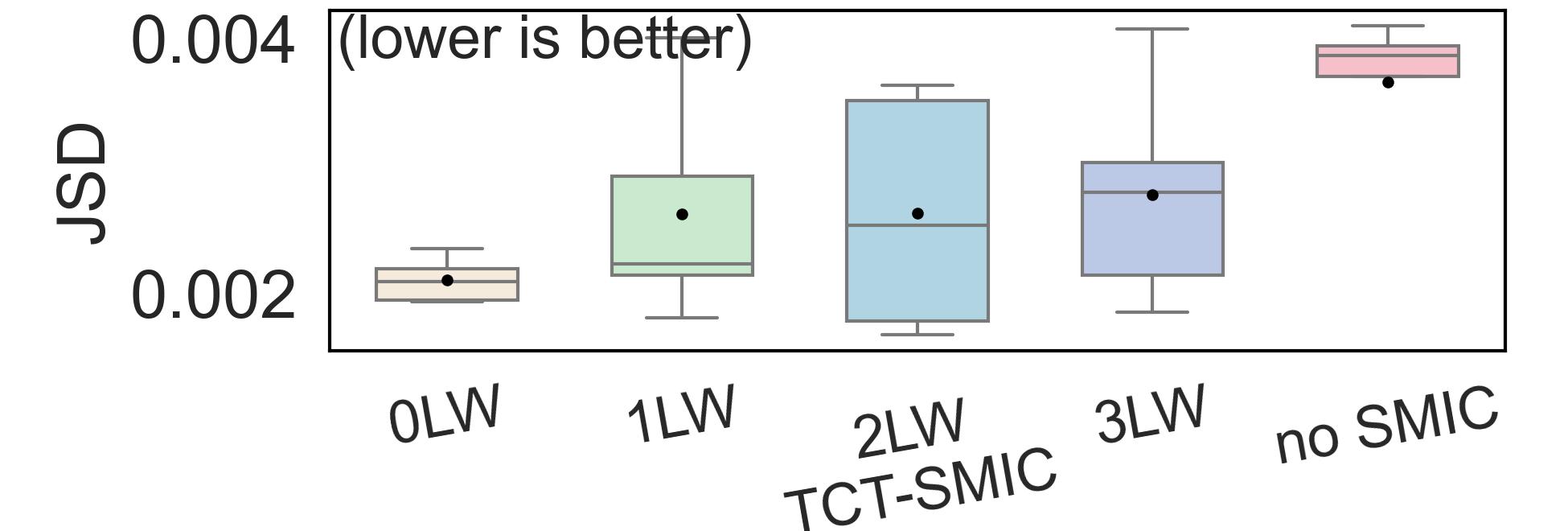}
	\caption{Attention diagnostics for SMIC ablations.}
	\label{1BR2}
\end{figure}

\begin{table}[t]
	\centering
	
	\begin{minipage}[t]{0.54\linewidth}
		\vspace{0pt}
		\centering
		\caption{Mean ANLI accuracy with and without SMIC.}
		\label{tab:ablation}
		\small
		\renewcommand{\arraystretch}{1.05}
		\setlength{\tabcolsep}{2pt}
		\begin{tabular*}{\linewidth}{
				@{\extracolsep{\fill}}lccc@{}
			}
			\toprule
			\textbf{Round} &
			\textbf{TCT-SMIC} &
			\textbf{w/o SMIC} &
			\textbf{Diff [95\% CI]} \\
			\midrule
			R1 & 39.03 & 40.22 & $-1.19$ $[-2.67,0.29]$ \\
			R2 & 44.89 & 43.72 & $1.17$ $[0.07,2.27]$ \\
			R3 & 43.17 & 41.78 & $1.39$ $[0.62,2.16]$ \\
			\bottomrule
		\end{tabular*}
		
		\par\smallskip
		{\footnotesize\raggedright
			*Accuracy is reported in percent.
			\par}
	\end{minipage}
	\hfill
	\begin{minipage}[t]{0.44\linewidth}
		\vspace{0pt}
		\centering
		\caption{Mean ANLI-R3 accuracy by number of learnable scales.}
		\label{tab:r3}
		\small
		\renewcommand{\arraystretch}{1.05}
		\setlength{\tabcolsep}{2pt}
		\begin{tabular*}{\linewidth}{
				@{\extracolsep{\fill}}lcc@{}
			}
			\toprule
			\textbf{Config} &
			\textbf{Acc} &
			\textbf{Diff [95\% CI]} \\
			\midrule
			2LW & 43.17 & --- \\
			1LW & 42.47 & $0.70$ $[-0.57,1.97]$ \\
			3LW & 41.57 & $1.60$ $[0.37,2.83]$ \\
			\bottomrule
		\end{tabular*}
	\end{minipage}
\end{table}
\paragraph{The number of learnable windows.}
Starting from the default \emph{2LW} setting, we vary the number of learnable scales (\emph{0LW/1LW/2LW/3LW}) across all \emph{SMIC} layers, leaving other components unchanged. In particular, \emph{0LW} disables the two Gaussian-mixture branches and retains the token-local and global branches.

On ANLI-R1, Fig.~\ref{1BR2} indicates that the fixed-scale configuration produces the lowest attention JSD but also the lowest deletion/insertion summary score. The best validation checkpoint is reached 1.7 epochs later on average than with the default two-scale configuration across 10 seeds. Fixed scales therefore slow convergence and change both attention diagnostics.

Figure~\ref{1BR2} (left) shows that moving from \emph{1LW} to \emph{2LW} improves deletion/insertion AUC, whereas \emph{3LW} is not consistently better than \emph{2LW} and has greater variability. 

On ANLI-R3, Table~\ref{tab:r3} indicates that the two-scale configuration has the highest observed mean accuracy at 43.17\%. The one-scale configuration reaches 42.47\%, with a confidence interval for the difference that includes zero. The three-scale configuration reaches 41.57\%, and the confidence interval for the difference excludes zero. Among the evaluated settings, two learnable receptive-field scales give the best balance of mean accuracy and attention diagnostics.

\paragraph{Contribution of \emph{SMIC}.}
Table~\ref{tab:ablation} shows that the effect of SMIC depends on the round. On ANLI-R1, the mean accuracy is 1.19 percentage points higher without SMIC, and the confidence interval includes zero. On ANLI-R2 and ANLI-R3, TCT-SMIC is higher by 1.17 and 1.39 percentage points. These two rounds also have the most training data, with 45,460 and 100,459 examples compared with 16,946 for ANLI-R1 \citep{nie2019adversarial}. The attention diagnostics show higher redistribution after SMIC is removed. On ANLI-R1, TCT-SMIC is not more accurate than the model without SMIC, so its lower redistribution does not come from higher accuracy.

\paragraph{Reference construction.}
Replacing the multiscale reference ${Z}^{(0)}$ with a token-level reference computed directly from token embeddings lowers mean ANLI-R2 accuracy from $44.89\%$ to $43.77\%$, a difference of $1.12$ percentage points ($95\%$ CI $[0.17,2.07]$). The multiscale construction therefore adds accuracy beyond a token-level input reference. A layer-local variant extracts token-span features from each layer output and uses no shared reference. On the same ANLI-R1 seeds, it reaches $32.92\%$, whereas TCT-SMIC reaches $39.03\%$.

\begin{figure}[t]
	\centering
	\includegraphics[width=0.49\linewidth]{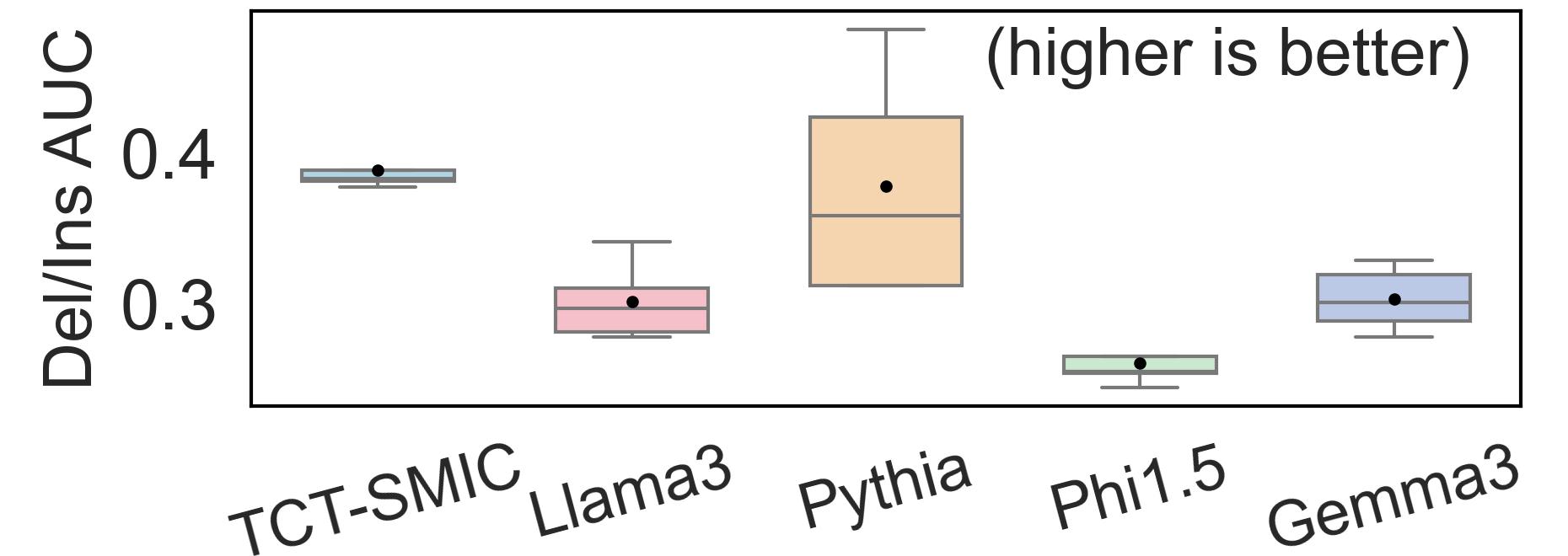}
	\hfill
	\includegraphics[width=0.49\linewidth]{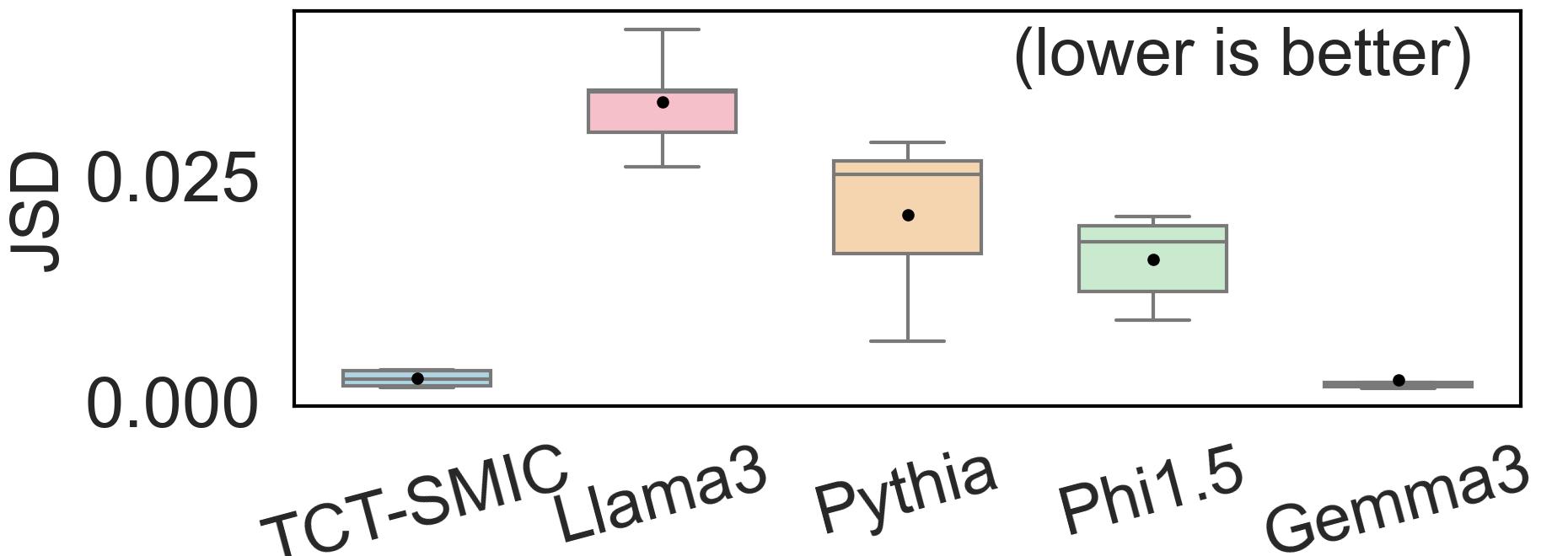}
	\caption{Attention diagnostics for three-layer models.}
	\label{fig:3llattn_drift}
\end{figure}

\paragraph{Depth-matched diagnostic.}
We compare the three-layer TCT-SMIC with three-layer Llama-3-style, Phi-1.5-style, Pythia-style, and Gemma-3-style models under the same training protocol. Figure~\ref{fig:3llattn_drift} shows that the higher deletion/insertion summary score and comparatively low attention redistribution persist in this depth-matched diagnostic. Thus, reduced depth alone does not explain the attention pattern.

\paragraph{Tokenizer sensitivity check.}
We rerun the ANLI-R1 seeds of TCT-SMIC with the Phi-1.5-style tokenizer. Mean accuracy is 39.03\% with the Llama-3-style tokenizer and 39.54\% with the Phi-1.5-style tokenizer. TCT-SMIC therefore retains comparable accuracy with a different tokenizer.

\section{Limitations}
\label{sec:conclusion}

First, the rule-generated surface-text perturbations are not manually re-annotated, and some, especially random replacements, may change the correct NLI relation.   The resulting accuracy differences therefore measure empirical sensitivity under this protocol, not robust accuracy with preserved labels.
  Second, because the baselines differ in architecture and tokenizer, Table~\ref{exp2} compares complete configurations and does not isolate SMIC. The within-architecture ablations and Llama-3-style plug-in provide more direct component-level evidence.
  Third, all accuracy comparisons use models trained from scratch for round-specific ANLI classification. The pretrained experiments cover two backbones and evaluate deletion/insertion AUC. Finally, deletion/insertion AUC and conditional pooled-attention JSD are descriptive diagnostics, not causal explanations or faithfulness guarantees.
  
  \section{Conclusion}\label{sec:final}
  We introduced SMIC, which builds one multiscale reference from the input and gives every Transformer layer direct access to it through a separate transformation per layer. Trained from scratch on the three ANLI rounds, TCT-SMIC reaches the highest ten-seed mean accuracy among decoder-based and MLP baselines of similar size and needs less training compute than most of them. The ablations further support the contribution of SMIC. Under sparse lexical replacements, TCT-SMIC shows comparatively low attention redistribution and the smallest mean signed accuracy change. Adding SMIC to a real-valued Llama-style model and to pretrained ModernBERT and Llama backbones raises the deletion/insertion score. The bound that links mean Jensen--Shannon divergence to attention redistribution holds for any model, so these diagnostics can compare other architectures on the same scale.
  
  \section*{Funding}
  This research did not receive any specific grant from funding agencies in the public, commercial, or not-for-profit sectors.
  
  \section*{Declaration of competing interest}
  The authors declare that they have no known competing financial interests or personal relationships that could have appeared to influence the work reported in this paper.
  
  \section*{Data availability}
  The ANLI dataset used in this study is publicly available \citep{nie2019adversarial}. The code is available from the corresponding author upon reasonable request.
  
\section*{Declaration of generative AI and AI-assisted technologies in the manuscript preparation process}
During the preparation of this work the authors used Claude to polish the writing and to search for related work. After using this tool, the authors reviewed and edited the content as needed, checked every suggested reference against the original publication, and take full responsibility for the content of the published article.


\appendix
\section{Derivation of the Cartesian form of the complex-vector token representation}
\label{app:complex-derivation}
Fix a token index $j\in\{1,\ldots,n\}$ and an embedding coordinate $k\in\{1,\ldots,d\}$. Define
$G_k=\sqrt{\sum_{v=1}^{n}w_{v,k}^{2}}.$
If $G_k=0$, then $w_{v,k}=0$ for every token position $v$. In this case, the numerically stabilized implementation returns
$Z_{j,k}=0.$
The exact derivation below therefore considers the nondegenerate case
$G_k>0.$
Define the adapted magnitude
$G_{j,k}'=G_k+w_{j,k},$
the normalized coordinate
$\tilde{w}_{j,k}=\frac{w_{j,k}}{G_k},$
and the phase
$\alpha_{j,k}=\operatorname{atan2}\left(\sqrt{1-\tilde{w}_{j,k}^{2}},\tilde{w}_{j,k}\right).$

In the implementation, denominators are lower-bounded by $\epsilon=10^{-8}$, squared cosine values are clipped to $[0,1]$, and the square-root argument is lower-bounded by $\epsilon$. The following equalities describe the corresponding idealized expression when these numerical safeguards are inactive.

Define the \emph{Complex-Vector Token Representation (CVTR)} in polar form as
$Z_{j,k}:=G'_{j,k}e^{\mathrm{i}\alpha_{j,k}}$.
Using Euler's formula $e^{\mathrm{i}\alpha}=\cos(\alpha)+\mathrm{i}\sin(\alpha)$, we obtain
$Z_{j,k}=G'_{j,k}\bigl(\cos(\alpha_{j,k})+\mathrm{i}\sin(\alpha_{j,k})\bigr)=T_{j,k}+\mathrm{i}C_{j,k}.$

\paragraph{Real part.}
$
T_{j,k}
=\Re(Z_{j,k})
=(G_k+w_{j,k})\cos(\alpha_{j,k})
=(G_k+w_{j,k})\frac{w_{j,k}}{G_k} 
=w_{j,k}+\frac{w_{j,k}^2}{G_k}
=w_{j,k}\left(1+\frac{w_{j,k}}{G_k}\right)
=(1+\cos(\alpha_{j,k}))\,w_{j,k}.
$

\paragraph{Imaginary part.}
$
C_{j,k}
=\Im(Z_{j,k})
=(G_k+w_{j,k})\sin(\alpha_{j,k})
=(G_k+w_{j,k})\frac{\sqrt{G_k^2-w_{j,k}^2}}{G_k}
=\left(1+\frac{w_{j,k}}{G_k}\right)\sqrt{G_k^2-w_{j,k}^2}
=(1+\cos(\alpha_{j,k}))\,\sqrt{G_k^2-w_{j,k}^2}.
$
Finally, since $G_k^2=\sum_{v=1}^n w_{v,k}^2$, we have
$G_k^2-w_{j,k}^2=\sum_{v\ne j} w_{v,k}^2$, and therefore
$C_{j,k}=(1+\cos(\alpha_{j,k}))\,\sqrt{\sum_{v\ne j} w_{v,k}^2}.$

\section{Proof of Proposition~\ref{prop:js_subset_mass}} \label{prop:js_subset_mass1}

\begin{proof}
	For two probability distributions $\mathbf{p}$ and $\mathbf{q}$ on the same finite support, define their total variation distance by
	$
	\operatorname{TV}(\mathbf{p},\mathbf{q})
	=
	\frac{1}{2}
	\|\mathbf{p}-\mathbf{q}\|_1.
	$
	
	For every subset $\mathcal{S}$ of the common support,
	$
	|p(\mathcal{S})-q(\mathcal{S})|
	\le
	\operatorname{TV}(\mathbf{p},\mathbf{q}).
	$
	Let
	$
	\mathbf{m}
	=
	\frac{\mathbf{p}+\mathbf{q}}{2}.
	$
	
	Pinsker's inequality gives
	$
	\mathrm{KL}(\mathbf{p}\|\mathbf{m})
	\ge
	2\operatorname{TV}(\mathbf{p},\mathbf{m})^2
	$
	and
	$
	\mathrm{KL}(\mathbf{q}\|\mathbf{m})
	\ge
	2\operatorname{TV}(\mathbf{q},\mathbf{m})^2.
	$
	Because
	$
	\operatorname{TV}(\mathbf{p},\mathbf{m})
	=
	\operatorname{TV}(\mathbf{q},\mathbf{m})
	=
	\frac{1}{2}
	\operatorname{TV}(\mathbf{p},\mathbf{q}),
	$
	the Jensen--Shannon divergence satisfies
	$
	J(\mathbf{p},\mathbf{q})
	\ge
	\frac{1}{2}
	\operatorname{TV}(\mathbf{p},\mathbf{q})^2.
	$
	It follows that
	$
	|p(\mathcal{S})-q(\mathcal{S})|
	\le
	\sqrt{2J(\mathbf{p},\mathbf{q})}.
	$
	
	Applying this inequality to every evaluated pair gives
	$
	\frac{1}{N}
	\sum_{i=1}^{N}
	\left|
	p_i(\mathcal{S}_i)
	-
	q_i(\mathcal{S}_i)
	\right|
	\le
	\frac{1}{N}
	\sum_{i=1}^{N}
	\sqrt{2J_i}.
	$
	Since the square-root function is concave, Jensen's inequality gives
	$
	\frac{1}{N}
	\sum_{i=1}^{N}
	\sqrt{2J_i}
	\le
	\sqrt{
		\frac{2}{N}
		\sum_{i=1}^{N}
		J_i
	}
	=
	\sqrt{2\overline{J}}.
	$
	Since $p_i(\mathcal{S}_i)$ and $q_i(\mathcal{S}_i)$ both lie in $[0,1]$, we also have
	$
	\left|
	p_i(\mathcal{S}_i)
	-
	q_i(\mathcal{S}_i)
	\right|
	\le
	1
	$
	for every $i$. Averaging over the evaluated pairs gives
	$
	\frac{1}{N}
	\sum_{i=1}^{N}
	\left|
	p_i(\mathcal{S}_i)
	-
	q_i(\mathcal{S}_i)
	\right|
	\le
	1.
	$
	Combining this bound with the Jensen--Shannon bound yields
	$
	\frac{1}{N}
	\sum_{i=1}^{N}
	\left|
	p_i(\mathcal{S}_i)
	-
	q_i(\mathcal{S}_i)
	\right|
	\le
	\min
	\left\{
	1,
	\sqrt{
		2\overline{J}
	}
	\right\},
	$
	which proves the proposition.
\end{proof}

\begin{remark}
	If Jensen--Shannon divergence is computed with base-$2$ logarithms, the right-hand side of Proposition~\ref{prop:js_subset_mass} becomes
	$
	\min
	\left\{
	1,
	\sqrt{
		2\ln(2)\overline{J}
	}
	\right\}.
	$
\end{remark}
\section{Implementation details}\label{app:settings}
Table~\ref{tab:settings} lists the settings of TCT-SMIC. The learning-rate search on ANLI-R1 covers $\{10^{-5},5\times10^{-5},8\times10^{-5},10^{-4},3\times10^{-4}\}$, and the dropout search covers $\{0,0.01,0.05,0.1\}$. The residual and feed-forward dropout of TCT-SMIC and the dropout of the MLP baseline are tuned separately and set to $0.175$ and $0.2$, respectively.

\begin{table}[h]
	\centering
	\caption{Implementation settings of TCT-SMIC.}
	\label{tab:settings}
	\small
	\begin{tabular}{@{}p{0.40\linewidth}p{0.56\linewidth}@{}}
		\toprule
		\textbf{Setting} & \textbf{Value} \\
		\midrule
		Attention layers, heads, feed-forward width & $3$, $32$, $4096$ \\
		Hidden size $d$ & $1024$ \\
		Tokenizer, maximum premise and hypothesis lengths & Llama-3.2-1B, $110$, $40$ \\
		Premise and hypothesis kernel sets & $\{1,3,7,13,21,33\}$, $\{1,3,5,7,9,13\}$ \\
		Layer-module kernel set & $\{1,3,7,13,21,33\}$ \\
		Anchor ratios (premise, hypothesis, layer modules) & $(0.17,0.33)$, $(0.30,0.65)$, $(0.17,0.33)$ \\
		Offset initialization, Gaussian bandwidth & $\mathcal{N}(0,0.5^{2})$, $\sigma=5$ \\
		Running-length update weights & $0.95$, $0.05$ \\
		Dropout (residual and feed-forward, attention) & $0.175$, $0.05$ \\
		Optimizer & AdamW \\
		Weight decay & $0.05$, and $0$ for biases and other one-dimensional parameters \\
		Learning rate & linear warmup from $4\times10^{-6}$ to $8\times10^{-5}$ over $3$ epochs, then cosine decay to $2\times10^{-5}$ at epoch $30$ \\
		Receptive-field offsets & $5$ times the learning rate, weight decay $0.01$ \\
		Gradient-norm clipping & $1.2$ \\
		Batch size & $16$ \\
		Maximum epochs, early-stopping patience & $30$, $5$ \\
		Training loss & cross-entropy of one shared classifier, averaged over the outputs of all layers, with predictions from the last layer \\
		Seeds & $10$ per round \\
		Precision & FP32, deterministic CUDA and cuDNN, TF32 disabled \\
		\bottomrule
	\end{tabular}
\end{table}

The Phi-1.5-style, Llama-3-style, Pythia-style, and Gemma-3-style baselines use the original model configurations, with $15$, $11$, and $12$ layers in the 0.8B Phi-1.5-style, Llama-3-style, and Pythia-style models and the full number of layers in the 1B group. They are trained from random initialization in FP32 with AdamW, peak learning rate $10^{-4}$, weight decay $0.1$, gradient-norm clipping at $1.0$, batch size $16$, at most $30$ epochs, and early stopping with patience $5$. The AdamW coefficients are $(0.9,0.98)$ for the Phi-1.5-style and Gemma-3-style models, $(0.9,0.95)$ for the Pythia-style models, and $(0.9,0.999)$ for the Llama-3-style models. The Llama-3-style models warm up linearly over the first $1\%$ of steps, follow a cosine decay to $1\%$ of the peak learning rate at $99\%$ of the steps, and then decay linearly to zero. The Pythia-style models warm up over the first $1\%$ of steps and follow a cosine decay to $10\%$ of the peak. The Phi-1.5-style and Gemma-3-style models keep the learning rate constant. Each of these four baselines truncates the premise and the hypothesis to $110$ and $40$ tokens as in TCT-SMIC and joins them into one sequence of at most $160$ tokens. Their learning rate and dropout are also tuned on ANLI-R1, and the selected dropout is each model's default. The MLP baseline applies $40$ fully connected layers of width $4096$ to the mean-pooled premise and hypothesis embeddings, their absolute difference, and their elementwise product. It uses dropout $0.2$ and the tokenizer, maximum input lengths, optimizer, and training loss of TCT-SMIC. Its tuned learning rate equals that of TCT-SMIC.

For the pretrained experiments, we keep the first three layers of Llama-3.2-1B and ModernBERT-large and fine-tune all parameters on ANLI-R1 with five seeds, batch size $16$, and at most $30$ epochs. The Llama backbone uses AdamW with peak learning rate $8\times10^{-5}$, weight decay $0.1$, attention dropout $0$, the learning-rate schedule of the Llama-3-style baselines, and early stopping with patience $5$. The ModernBERT backbone uses StableAdamW with peak learning rate $5\times10^{-5}$, coefficients $(0.9,0.98)$, epsilon $10^{-6}$, weight decay $10^{-5}$, dropout $0$, a warmup--stable--decay schedule with $1\%$ warmup and a final $3\%$ decay to zero, and early stopping with patience $7$. Each backbone uses the same settings with and without SMIC.

Training FLOPs in Table~\ref{exp2} are estimated as six times the number of non-embedding parameters times the number of training tokens \citep{kaplan2020scaling}. For every configuration, the number of training tokens is approximated by $2\times10^{6}$, $4\times10^{6}$, and $8\times10^{6}$ tokens per epoch on ANLI-R1, ANLI-R2, and ANLI-R3 and multiplied by the number of entries in the training logs of the ten seeds.
\printcredits

\bibliographystyle{cas-model2-names}

\bibliography{cas-refs}
\end{document}

%% file: math_commands.tex
\usepackage{amsmath,amsfonts,bm}

\def\eqref#1{equation~\ref{#1}}

\def\1{\bm{1}}

\DeclareMathAlphabet{\mathsfit}{\encodingdefault}{\sfdefault}{m}{sl}
\SetMathAlphabet{\mathsfit}{bold}{\encodingdefault}{\sfdefault}{bx}{n}

